\documentclass{article}

\usepackage{arxiv}

\usepackage[utf8]{inputenc} 
\usepackage[T1]{fontenc}    
\usepackage{hyperref}       
\usepackage{url}            
\usepackage{booktabs}       
\usepackage{amsfonts}       
\usepackage{nicefrac}       
\usepackage{microtype}      
\usepackage{lipsum}		
\usepackage{graphicx}
\usepackage{natbib}
\usepackage{doi}

\usepackage{algorithm}
\usepackage{algpseudocode}
\usepackage{amsmath,amsfonts}
\usepackage{cleveref}
\usepackage{adjustbox}
\usepackage{multirow}
\usepackage[table]{xcolor}
\usepackage{caption}
\usepackage{subcaption}
\usepackage{booktabs}
\newcommand{\mathE}{\mathbb{E}}

\DeclareMathOperator*{\argmin}{arg\,min}

\newcommand{\eg}{\textit{e.g\@.}}

\newcommand{\etal}{\textit{et~al\@.}}

\newcommand{\vs}{\textit{vs\@.}}

\newcommand{\nas}{H-Meta-NAS}

\title{Rapid Model Architecture Adaption for Meta-Learning}

\author{
	Yiren Zhao \\
	University of Cambridge \\
	yaz21@cam.ac.uk \\
	\And
	Xitong Gao \\
	Shenzhen Institute of Advanced Technology, CAS \\
	xt.gao@siat.ac.cn \\
	\And
	Ilia Shumailov \\
	University of Cambridge \\
	is410@cam.ac.uk \\
	\And
	Nicolo Fusi \\
	Microsoft Research \\
	fusi@microsoft.com \\
	\And
	Robert Mullins \\
	University of Cambridge \\
	Robert.Mullins@cl.cam.ac.uk \\

}

\date{}

\renewcommand{\shorttitle}{Rapid Model Architecture Adaption for Meta-Learning}

\hypersetup{
pdftitle={A template for the arxiv style},
pdfsubject={q-bio.NC, q-bio.QM},
pdfauthor={David S.~Hippocampus, Elias D.~Striatum},
pdfkeywords={First keyword, Second keyword, More},
}

\begin{document}
\maketitle

\begin{abstract}
	Network Architecture Search (NAS) methods have recently gathered much attention.
	They design networks with better performance 
	and use a much shorter search time compared to traditional manual tuning.
	Despite their efficiency in model deployments, 
	most NAS algorithms target a single task on a fixed hardware system.
	However, real-life few-shot learning environments often cover a great number of tasks ($T$) 
	and deployments on a wide variety of hardware platforms ($H$). 
	
	The combinatorial search complexity $T \times H$ creates
	a fundamental search efficiency challenge if one naively applies existing NAS methods to these scenarios.
	To overcome this issue, we show, for the first time, how to rapidly adapt model architectures to new tasks
	in a \emph{many-task many-hardware} few-shot learning setup by integrating Model Agnostic Meta Learning (MAML) 
	into the NAS flow. 
	The proposed NAS method (H-Meta-NAS)
	is hardware-aware and performs optimisation in the MAML framework.
	\nas~shows a Pareto dominance compared to a variety of NAS and manual baselines in popular few-shot learning benchmarks with various hardware platforms and constraints.
	In particular, on the 5-way 1-shot Mini-ImageNet classification task, 
	the proposed method 
	outperforms the best manual baseline
	by a large margin ($5.21\%$ in accuracy) using $60\%$ less computation.
\end{abstract}


\section{Introduction}

\begin{figure}[]
	\begin{center}
        \includegraphics[width=.8\linewidth]{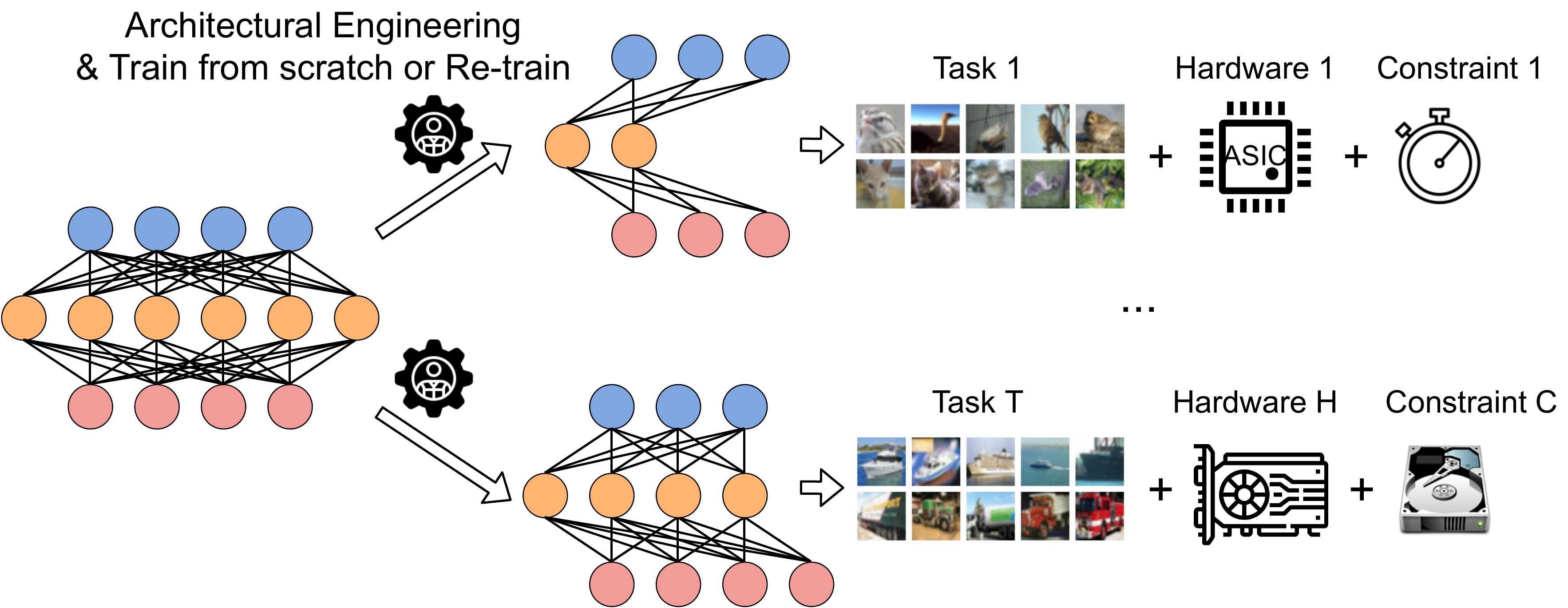}
	\end{center}
	\caption{
        Deploying networks in a \emph{many-task many-device} few-shot learning setup. This implies a large search complexity $\mathcal{O}(THC)$.
		}
	\label{fig:motivation}
    \vspace{-15pt}
\end{figure}

Existing Network Architecture Search (NAS) methods show promising performance on image \citep{zoph2016neural, liu2018darts}, language \citep{guo2019nat,so2019evolved} and graph data \citep{zhao2020probabilistic}.
The automation not only reduces the human effort required for architecture tuning but also produces architectures with state-of-the-art performance in domains like image classification \citep{zoph2016neural} and language modeling \citep{so2019evolved}.
Most NAS methods today focus on a single task with a fixed hardware system, yet real-life model deployments covering multiple tasks and various hardware platforms will significantly prolong this process.
As illustrated in \Cref{fig:motivation}, a common design flow is to re-engineer the architecture and train for different task($T$)-hardware($H$) pairs with different constraints ($C$). 
The architectural engineering phase can be accomplished whether manually or by using an established NAS procedure.
The major challenge is designing an efficient algorithmic method to overcome the quickly scaling $\mathcal{O}(THC)$ search complexity described in \Cref{fig:motivation}. 

Few-shot learning systems follow exactly this \emph{many-task many-device} setup,
when considering deployments on different user devices on key applications 
such as facial~\citep{guo2020letafacerecog} and speech recognition~\citep{hsu2020meta}.
A task in few-shot learning normally takes an $N$-way $K$-shot
formulation, where it contains $N$ classes with $K$ support samples and $Q$ query samples in each class.
Model-Agnostic Meta-Learning (MAML), incorporating the idea of learning to learn, builds a meta-model using a great number of training tasks, and then adapts the meta-model to unseen test tasks using only a very small number of gradient updates \citep{finn2017model}.
MAML then becomes a powerful and elegant approach for few-shot learning -- its ability to quickly adapt to new tasks can potentially shrink the $\mathcal{O}(THC)$ complexity illustrated in \Cref{fig:motivation} to $\mathcal{O}(HC)$. 
In the meantime, hardware-aware NAS methods \citep{cai2019once, cai2018proxylessnas,xu2020latency}, \eg~the train-once-for-all technique~\citep{cai2019once}, support deployments of searched models to fit to different hardware platforms with various latency constraints. 
These hardware-aware NAS techniques further reduce the search complexity from $\mathcal{O}(THC)$ to $\mathcal{O}(T)$ \citep{cai2018proxylessnas}.

In this paper, we propose a novel Hardware-aware Meta Network Architecture Search (\nas).
Integration of the MAML framework into hardware-aware NAS theoretically reduces the search complexity from $\mathcal{O}(THC)$ to $\mathcal{O}{(1)}$, allowing for a rapid adaption of model architectures to unseen tasks on new hardware systems.
However, we identified the following challenges in this integration:
\begin{itemize}
    \item Classic NAS search space contains many over-parameterised sub-models, this makes it hard to tackle the over-fitting phenomenon in few-shot learning.
    \item Hardware-aware NAS profiles latency for sub-networks on each task-hardware pair, this profiling can be prolonged significantly with a great number of tasks and, more importantly, if the targeting device has scarce computation resources.
\end{itemize}

To tackle these challenges, we then propose to use Global Expansion (GE) and Adaptive Number of Layers (ANL) to allow a drastic change in model capabilities for tasks with varying difficulties. Our experiments later demonstrate that such changes alleviate over-fitting in few-shot learning and improve the accuracy significantly.
We also present a novel layer-wise profiling strategy to allow reuse of profiling information across different tasks.

In this paper, we make the following contributions:
\begin{itemize}
    \item We propose a novel Hardware-aware Network Architecture Search for Meta learning (H-Meta-NAS). H-Meta-NAS quickly adapts meta-architectures to new tasks with hardware-awareness and can be conditioned with various device-specific latency constraints. 
    The proposed NAS reduces search complexity from $\mathcal{O}(THC)$ to $\mathcal{O}(1)$ in a realistic \emph{many-task many-device} few-shot learning setup. We extensively evaluate H-Meta-NAS on various hardware platforms (GPU, CPU, mCPU, IoT, ASIC accelerator) and efficiency constraints (latency and model size), our latency-accuracy performance curve demonstrates a pareto dominance.
    \item We propose a task-agnostic layer-wise profiling strategy for the NAS. This profiling reduces the profiling run-time from around $10^5$ hours to 1.2 hours when targeting hardware with limited capabilities (\eg~IoT devices).
    \item We show several tricks for the NAS algorithm, named Global Expansion and Adaptively Number of Layers respectively. These methods help the NAS to overcome the over-fitting problem in few-shot learning from the architectural perspective.
\end{itemize}

\section{Related work}

\paragraph{Few-shot learning in the MAML framework}
Inspired by human's ability to learn from only a few tasks and generalise the knowledge to unseen problems, a meta learner is trained over a distribution of tasks with the hope of generalising its learned knowledge to new tasks \cite{finn2017model}.
\begin{equation}
    \argmin_{\theta} (\mathE_{\mathcal{T} \in \mathbb{T}}[\mathcal{L}_{\theta}(\mathcal{T})])
    \label{equ:meta}
\end{equation}
\Cref{equ:meta} captures the optimisation objective of meta-learning, 
where optimal parameters are obtained through optimising on a set of \textit{meta-training} tasks.
Current mainstream approaches of 
using meta-learning to solve few-shot learning problems 
can be roughly categorised into three types: Memory-based, Metric-based and Optimisation-based.

Memory-based method utilises a memory-augmented neural network \citep{munkhdalai2017meta, gidaris2018dynamic}
to memorise meta-knowledge for a fast adaption to new tasks.
Metric-based methods aim to meta-learn a high-dimensional feature representation of samples, and then apply certain metrics to distinguish them. For instance, Meta-Baseline utilises the cosine nearest-controid metric \citep{chen2020new} and DeepEMD applies the Wasserstein distance \citep{zhang2020deepemd}. Optimisation-based method, on the other hand, focuses on learning a good parameter initialisation (also known as \emph{meta-parameters} or \emph{meta-weights}) from a great number of training tasks, 
such that these meta-parameters adapt to new few-shot tasks within a few gradient updates. 
The most well-established Optimisation-based method is Model-Agnostic Meta-Learning (MAML) \citep{finn2017model}. MAML is a powerful yet simple method to tackle the few-shot learning problem, since its adaption relies solely on gradient updates. \citeauthor{antoniou2018train}~later demonstrate MAML++, a series of modifications that improved MAML's performance and stability. \citeauthor{NEURIPS2020_ee89223a} introduce an additional network for generating adaptive parameters for the inner-loop optimisation.

Despite the rise in popularity of the meta-learning framework applied to few-shot learning, 
little attention has been paid to the runtime efficiency of these approaches.
Meta-learning has been explored in key applications such as 
facial and speech recognition \citep{hsu2020meta,guo2020letafacerecog} for mobile deices.
Real-life deployments on these devices resemble a \emph{many-task many-device} scenario,
where learning on each user's data 
is a few-shot learning task 
and different hardware platforms represent different types of under-deployment devices. 
Memory-based and Metric-based meta-learning methods are then challenged by the hardware or latency constraints: 
Memory-based methods need additional storage space (at least double)
and Metric-based approaches use multiple inference runs (at least two) for a single image classification. 
In this work, we then focus solely on an Optimisation-based approach because of the runtime concern outlined above.
The proposed NAS method utilises the simple yet effective MAML++ framework: after adapting the model to new tasks, MAML++ executes exactly one inference run for a single test sample without additional memory usage.

\paragraph{Network architecture search}
Architecture engineering is a tedious and complex process requiring a lot of effort from human experts.
Network Architecture Search (NAS) focuses on 
reducing the amount of manual tuning in this design space. 
Early NAS methods use evolutionary algorithms and reinforcement learning to traverse the search space \citep{zoph2016neural,real2017large}.
These early methods require scoring architectures trained to a certain convergence and thus use a huge number of GPU hours.
Two major directions of NAS methods, Gradient-based and Evolution-based methods, 
are then explored in parallel in order to make the search cost more affordable.
Gradient-based NAS methods use Stochastic Gradient Descent (SGD) 
to optimise a set of probabilistic priors that are associated with architectural choices \citep{liu2018darts,casale2019probabilistic}. 
Although these probabilistic priors can be made latency-aware \citep{wu2019fbnet,xu2020latency}, 
it is challenging to make them follow a hard latency constraint.
Evolution-based NAS, on the other hand, 
operates on top of a pre-trained super-net and use evolutionary algorithms or reinforcement learning to 
pick best-suited sub-networks \citep{cai2018proxylessnas,cai2019once}, making it easier to be constrained by certain hardware metrics.
For instance, Once-for-all (OFA) is an Evolution-based NAS method and its searched networks are not only optimised for a specific hardware target but also constrained by a pre-defined latency budget~\citep{cai2019once}.
Our proposed \nas~shares certain similarities to Once-for-all, since this method offers a chance to reduce the hardware search complexity from $\mathcal{O}(HC)$ to $\mathcal{O}(1)$.

Several NAS methods are proposed under the MAML framework \citep{kim2018auto,shaw2018meta,lian2019towards}, 
these methods successfully reduce the search complexity from $\mathcal{O}(T)$ to $\mathcal{O}(1)$.
However, some of these methods do not show significant performance improvements compared to carefully designed MAML methods (\eg~MAML++) \citep{kim2018auto,shaw2018meta}.
In the meantime, some of these MAML-based NAS methods follow the Gradient-based approach and operate on complicated cell-based structures \citep{lian2019towards}. 
We illustrate later how cell-based NAS causes an undesirable effect on latency, and also meets fundamental scalability challenges when trying to deploy in a \emph{many-task many-device} few-shot learning setup.

\section{Method}


\begin{figure*}[!h]
	\centering
	\includegraphics[width=.8\linewidth]{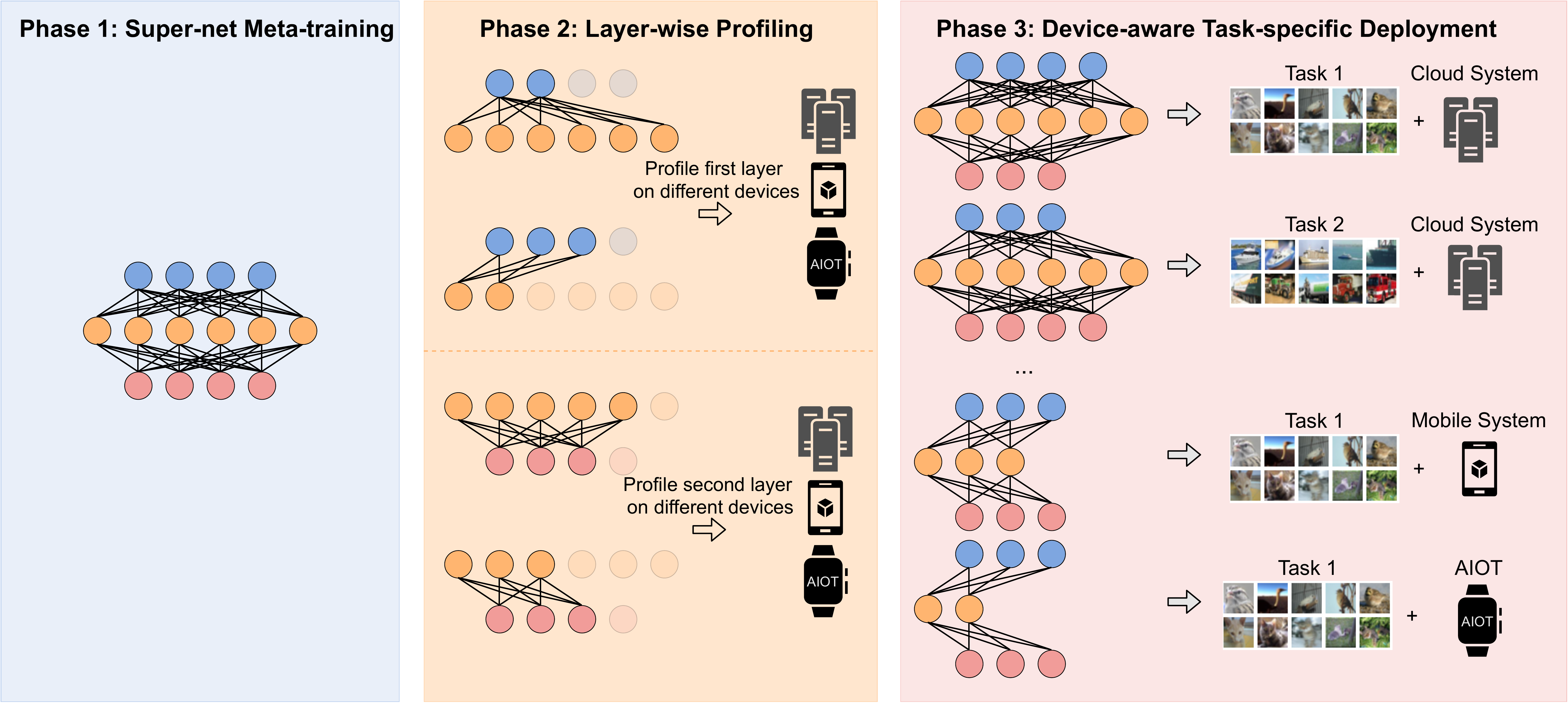}
	\caption{An overview of the three main stages of the proposed \nas~algorithm.}
	\label{fig:overview}
	\vspace{-15pt}
\end{figure*}

\textbf{Problem formulation}
In the MAML setup, we consider 
a set of tasks $\mathbb{T}$ and each task $\mathcal{T}_i \in \mathbb{T}$ 
contains a support set $\mathcal{D}_i^s$ and a query set $\mathcal{D}_i^q$.
The support set is used for task-level learning while the query set is in charge of 
evaluating the meta-model.
All tasks are divided into three sets, namely meta-training ($\mathbb{T}_{train}$), 
meta-validation ($\mathbb{T}_{val}$) and meta-testing ($\mathbb{T}_{test}$) sets.

\Cref{equ:pretrain} formally states the objective of the pre-training stage 
illustrated in \Cref{fig:overview} Phase 1.
The objective of this process is to optimise the parameters $\theta$ of the super-net
for various sub-networks sampled from the architecture set $\mathbb{A}$.
This will ensure the proposed \nas~to have both the meta-parameters 
and meta-architectures ready for the adaption to new tasks.
\begin{equation}
    \argmin_{\theta} \mathE_{\alpha \sim p(\mathbb{A})}[\mathE_{\mathcal{T} \in \mathbb{T}_{train}}[\mathcal{L}_{\theta}(\mathcal{T}, \alpha)]]
    \label{equ:pretrain}
\end{equation}
\Cref{equ:adapt} describes how \nas~adapts network architectures to a particular task $\mathcal{T}$ 
with a given hardware constraint $\mathit{C}_h$ (Phase 3 in \Cref{fig:overview}).
In practice, using the support set data $\mathcal{D}_i^s$ from a target task $\mathcal{T}_i$, 
we apply a genetic algorithm for finding the optimal architectures $\alpha^*$. 
We discuss further how this process in details in later sections.
\begin{equation}
	\begin{aligned}
		\alpha^* = \min_{\alpha} & \sum_{\alpha \in \mathbb{A}}\mathcal{L}_{\theta}(\mathcal{D}_i^s, \alpha) \\
		\textrm{s.t.} \quad & \mathcal{C}(\alpha) \leq \mathit{C}_h \\
		\label{equ:adapt}
	\end{aligned}
\end{equation}

\vspace{-15pt}
\paragraph{Architecture space}
\label{sec:method:searchspace}
\nas~considers a search space composed of different kernel sizes,
number of channels and activation types.
We mostly consider a VGG9-based NAS backbone, 
that is a 5-layer CNN model with the last layer being a fully connected layer.
We chose this NAS backbone because both MAML \cite{finn2017model} and MAML++ \cite{antoniou2018train}
used a VGG9 model architecture. The details of this backbone  are in Appendix.

We allow kernel sizes to be picked from $\{1, 3, 5\}$, channels to be expanded with a set of scaling factors $\{0.25, 0.5, 0.75, 1, 1.5, 2, 2.25\}$ and also six different activation functions (details in Appendix). For a single layer, there is $3\times 7 \times 6=126$ search options. \nas~also contains an Adaptive Number of Layers strategy, the network is allowed to use a subset of the total layers in the supernet with a maximum usage of $4$ layers.
The whole VGG9-based backbone then gives us in total $126^4 \times 4 \approx 10^9$ possible neural network architectures.

In addition, to demonstrate the ability of \nas~on more complex NAS backbone.
We also studied an alternative ResNet12-based NAS backbone, that has approximately $2 \times 10^{24}$ possible sub-networks.

\paragraph{Super-net meta-training strategy}
\label{sec:method:pretrain}
As illustrated by prior work \cite{cai2019once}, progressively shrinking the super-net during meta-training can reduce the interference between sub-networks.
We observe the same phenomenon and then use a similar progressive shrinking strategy in \nas, the architectural sampling process $\alpha \sim p(\mathbb{A})$ will pick the largest network with a probability of $p$, and randomly pick other sub-networks with a probability of $1-p$.
We apply an exponentially decay strategy to $p$:

\begin{equation}
	p = p_e + (p_i - p_e) \times exp(- \alpha \times \frac{e-e_s}{e_{m} - e_s}))
\end{equation}
$p_e$ and $p_i$ are the end and initial probabilities. $e$ is the current number of epochs, and $e_s$ and $e_m$ are the starting and end epochs of applying this decaying process. $\alpha$ determines how fast the decay is.
In our experiment, we pick $p_i=1.0$ and $e_s=30$, because the super-net reaches a relatively stable training accuracy at that point.
We then start the decaying process, and the value $\alpha=5$ is determined through a hyper-parameter study shown in our Appendix.

\paragraph{Layer-wise profiling}
\label{sec:method:profile}
Hardware-aware NAS needs the run-time of sub-networks on the targeting hardware to guide the search process \cite{cai2019once,xu2020latency}.
However, the profiling stage can be time-consuming if given a low-end hardware as the profiling target and the search space is large.
For instance, running a single network inference of VGG9 on the Raspberry Pi Zero with a 1GHz single-core ARMv6 CPU takes around $2.365$ seconds to finish.
If we assume this is the averaged time needed for profiling a sub-network,
given that the entire search space includes around $10^9$ sub-networks, a naive traverse will take a formidable amount of time which is approximately $6 \times 10^5$ hours.
More importantly, the amount of profiling time scales with the number of hardware devices ($\mathcal{O}(H)$). 
Existing hardware-aware NAS schemes build predictive methods to estimate the run-time of sub-networks \cite{cai2019once,xu2020latency} and have a relatively significant error. We show in our evaluation, performing an exact profiling can be done with a low cost if allowing a per-layer profiling strategy.  


\paragraph{Adaption strategy}
\label{sec:method:adapt}
The adaption strategy uses a genetic algorithm \cite{whitley1994genetic} to pick the best suited sub-network with respective to a given hardware constraint, the full algorithm is detailed in Appendix.
In general, the adaption algorithm randomly samples a set of tasks from $\mathbb{T}_{val}$, and uses the averaged loss value and satisfaction to the hardware constraints as indicators the for the genetic algorithm.
The genetic algorithm has a pool size $P$ and number of iterations $M$, we demonstrate the optimal values are $P=100, M=200$ in our evaluation.


\paragraph{NAS backbone design}
One particular problem in few-shot learning is that models are prone to over-fitting.
This is because only a small number of training samples are available for each task and the network normally iterate on these sample many times.
We would like to explore on the architectural space to help models to overcome over-fitting and conduct a case study for different design options available for the backbone network.
We identify the following key changes to the NAS backbone to help the models to have high accuracy in few-shot learning:

\begin{itemize}
    \item $n \times n$ pooling: Pooling that applied to the final convolutional operation, $n \times n$ indicates the height and width of feature maps after pooling.
    \item Global Expansion (GE): Allowing the NAS to globally expand or shrink the number of channels of all layers.
    \item Adaptive Number of Layers (ANL): Allowing the NAS to use an arbitrary number of layers, the network then is able to early stop using only a fewer number of layers.
\end{itemize}

\begin{figure*}[!t]
	\centering
	\includegraphics[width=0.8\linewidth]{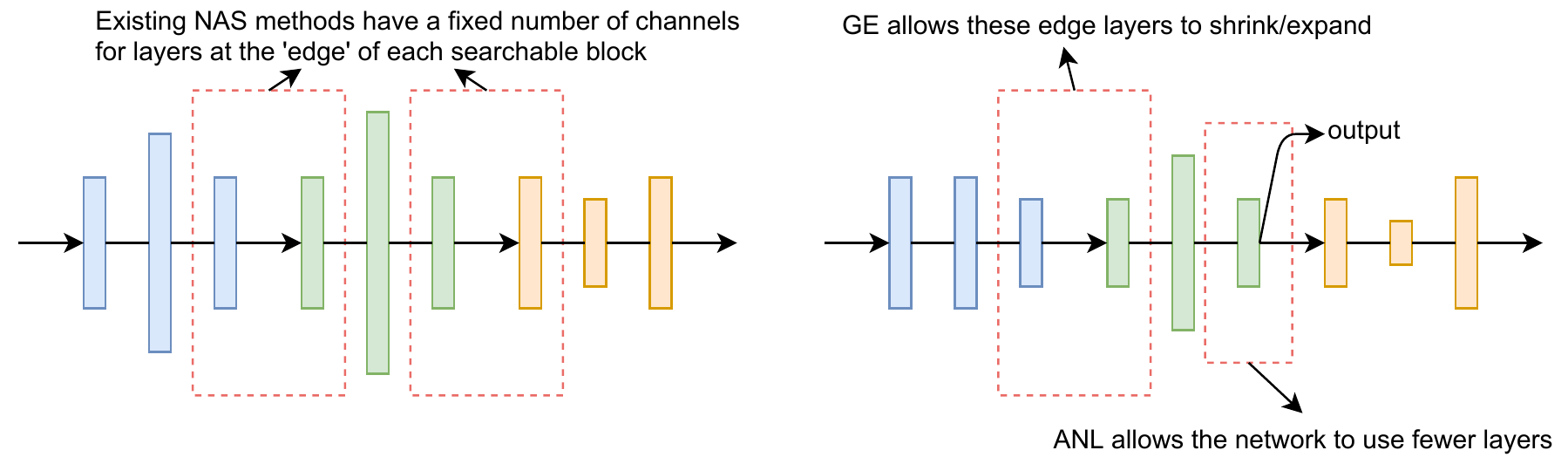}
	\caption{A graphical illustration of GE and ANL. Both methods will allow a more drastic change in model capabilities, allowing the searched model to deal with tasks with varying difficulties.}
	\label{fig:tricks}
	\vspace{-10pt}
\end{figure*}

\Cref{fig:tricks} further illustrate that GE and ANL can allow a much smaller model compared to existing NAS backbones.
We then demonstrate using a case study in our evaluation how a combination of these techniques can help \nas: the final searched model can have an up to $14.28\%$ accuracy increase on the 5-way 1-shot Mini-ImageNet classification if using these optimisation tricks.

\section{Evaluation}



\begin{table}[]
  \caption{
    Details of hardware systems experimented with \nas.}
  \label{tab:hardware}
  \begin{center}
  \adjustbox{scale=0.7}{%
  \begin{tabular}{cc}
    \toprule
    System
    & Device 
    \\
    \midrule
    Cloud
    & Nvidia GeForce RTX 2080 Ti
    \\
    Mid-end CPU 
    & Intel CPU
    \\
    Mobile CPU
    & Raspberry Pi 4B
    \\
    IoT
    & Raspberry Pi Zero 
    \\
    ASIC
    & Eyeriss \cite{chen2016eyeriss}
    \\
    \bottomrule
    \end{tabular}
}
  \end{center}
\vspace{-15pt}
\end{table}

We evaluate \nas~in a few-shot learning setup.
For each dataset, we search for the meta-architecture and meta-parameters.
We then adapt the meta-architecture with respect to a target hardware-constraint pair.
In the evaluation stage, we then re-train the obtained hardware-aware task-specific architecture to convergence and report the final accuracy.
We consider three popular datasets in the few-shot learning community: Omniglot, Mini-ImageNet and Few-shot CIFAR100. We use the PytorchMeta framework to handle the datasets \citep{deleu2019torchmeta}.

\textbf{Omniglot} is a handwritten digits recognition task, containing 1623 samples \citep{lake2015human}. 
We use the meta train/validation/test splits used Vinyals \etal~\cite{vinyals2016matching}. These splits are over 1028/172/423 classes (characters).

\textbf{Mini-ImageNet} is first introduced by Vinyals \etal. 
This dataset contains images of 100 different classes from the ILSVRC-12 dataset \citep{deng2009imagenet}, the splits are taken from Ravi \etal \citep{ravi2016optimization}.

\textbf{FC100} is introduced by \citeauthor{oreshkin2018tadam}, the datasets has 100 different classes from the CIFAR100 dataset \cite{krizhevsky09learningmultiple}.

\Cref{tab:hardware} details the systems and representative devices considered. Our Appendix contains a more detailed explanation of the specs of each hardware device. We use the ScaleSIM cycle-accurate simulator \cite{samajdar2018scale} for the Eyeriss \cite{chen2016eyeriss} accelerator. Details about this simulation and more information with respect to the datasets and search configurations are in our Appendix. 

\paragraph{The effect of pool sizes}
\begin{figure}[]
	\centering
	\includegraphics[width=.5\linewidth]{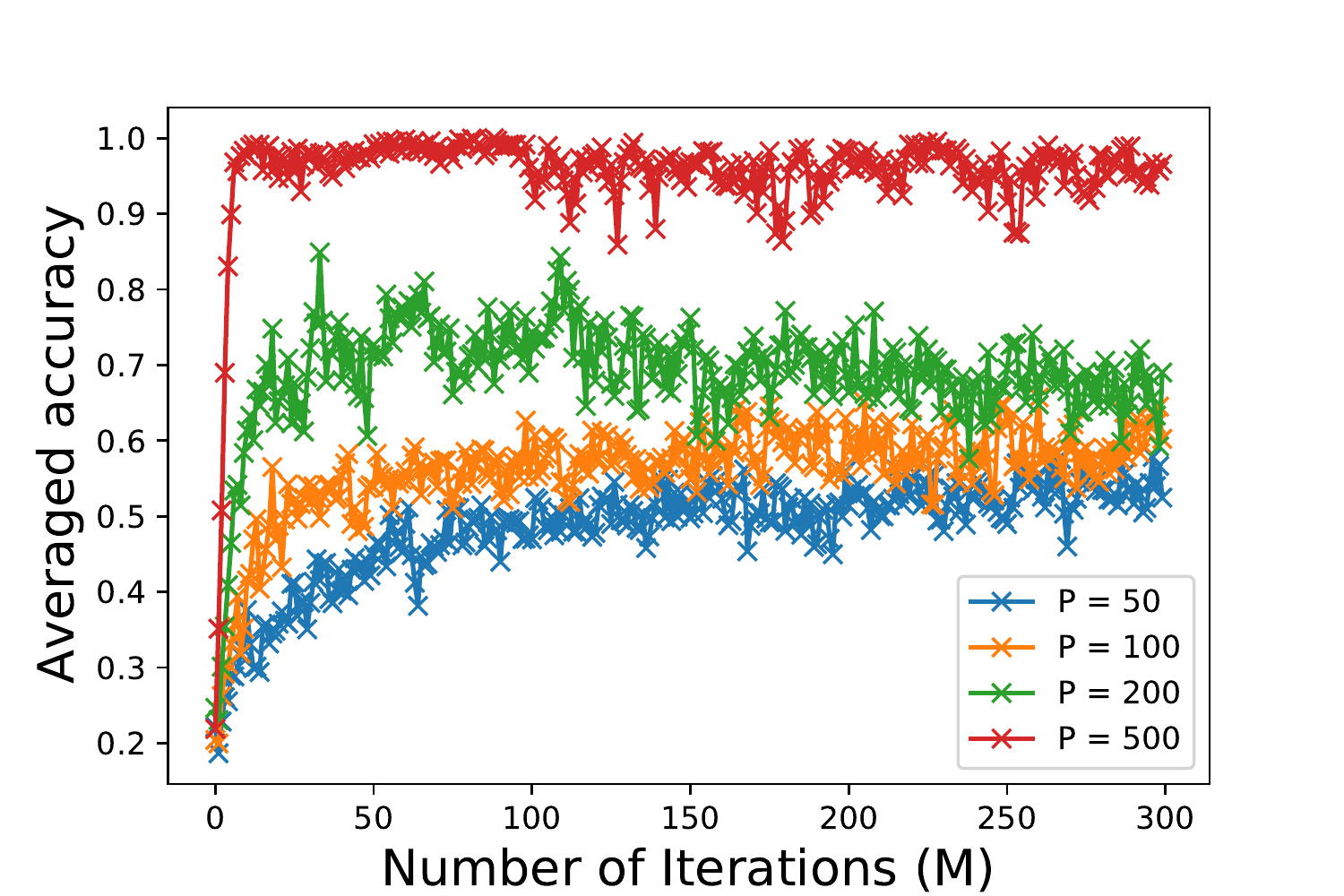}
	\caption{The effect of different pool size (P) and different number of iterations (M).}
	\label{fig:pool}
  \vspace{-15pt}
\end{figure}
We identify the following two hyper-parameters that can potentially affect the performance in the adaption stage, namely the number of iterations $M$ and the pool size $P$, and then run an hyper-parameter analysis in \Cref{fig:pool}. The full adaption algorithm making use of these hyper-parameters is in our Appendix. The horizontal axis shows the number of iterations and the vertical axis shows the averaged accuracy on the sampled tasks for all architectures in the pool.
\Cref{fig:pool} shows that the accuracy convergence is reached after around $150$ iterations, and running for additional iterations only provides marginal accuracy gains.
For this reason, we picked the number of iterations to be $200$ for a balance between accuracy and run-time.
In the meantime, we notice in general a higher pool size will give better adapted accuracy. However, 
this does not mean the final searched accuracy is affected to the same degree.
The final re-trained accuracies of searched architectures show an accuracy gap of $0.21\%$ between $P=100$ and $P=200$ and $0.32\%$ between $P=100$ and $P=500$.
An increase in pool size can prolong the run-time significantly, we thus picked a pool size of $100$ since it offers the best balance between accuracy and run-time.

\begin{table}[]
    \caption{
        A case study of different design options for the NAS backbone network.
        Experiments are executed with a model size constraint of $70K$
        on the Mini-ImageNet 5-way 1-shot classification task.
    }
    \label{tab:case_study}
    \begin{center}
    \adjustbox{scale=0.8}{
    \begin{tabular}{c|c}
    \toprule
    Design options & Accuracy \\
    \midrule
    MAML
    & $48.70\%$   \\
    MAML++ 
    & $52.15\%$   \\
    \midrule
    \nas~$+$ $1\times1$ Pool & $42.28\%$ \\
    \nas~$+$ $5\times5$ Pool & $46.13\%$ \\ 
    \nas~$+$ $5\times5$ Pool $+$ GE
    & $53.09\%$ \\
    \nas~$+$ $5\times5$ Pool $+$ GE $+$ ANL
    & $56.35\%$ 
    \\
    \bottomrule
    \end{tabular}
    }
    \end{center}
\end{table}
\paragraph{Evaluating pooling, GE and ANL}
Our results in \Cref{tab:case_study} suggest that a correct pooling strategy, GE and ANL can change the NAS backbone to allow the search space to reach much smaller models and thus provide a better accuracy.
In addition, \Cref{tab:case_study} also illustrates that $5 \times 5$ pooling is necessary for a higher accuracy.
We hypothesize this is because a relatively large fully-connected layer after the pooling is required for the network to achieve a good accuracy in this few-learning setup.

\paragraph{Latency predictor \vs~layer-wise latency profiling}
\begin{table}
    \caption{
        Comparing latency predictor with our proposed profiling.
        MSE Error is the error between estimated and measured latency, 
        Time is the total time taken to collect and build the estimator.
    }
    \label{tab:latency}
    \begin{center}
    \adjustbox{scale=0.7}{%
    \begin{tabular}{c|c|cc}
    \toprule

    Hardware
    & Metric
    & Latency Predictor
    & Layer-wise Profiling \\
    \midrule

    \multirow{2}*{2080 Ti GPU}
    & MSE Error
    & 0.0188 
    & 0.00690 \\
    & Time
    & 16.09 mins
    & 6.216 secs \\
    \midrule

    \multirow{2}*{Intel i9 CPU}
    & MSE Error
    & 0.165
    & 0.0119 \\
    & Time
    & 21.92 mins
    & 16.41 secs \\
    \midrule

    \multirow{2}*{Pi Zero}
    & MSE Error
    & N/A
    & 0.00742 	 \\
    & Time
    & N/A (Approx. 220 hours)
    & 82.41 mins \\

    \bottomrule
    \end{tabular}
    }
    \end{center}
\end{table}
We re-implemented the latency predictor in OFA \cite{cai2019once} to illustrate how a layer-wise profiling and look-up method is a perfect match in our learning scenario. We pick 16K training samples and 10K validation samples to train and test the latency predictor, which is the same as setup used in OFA. We use another 10K testing samples to evaluate the performance of OFA-based latency predictor against our layer-wise profiling on different hardware systems in terms of MSE (measuring the latency estimation quality) and Time (measuring the efficiency).

As illustrated in \Cref{tab:latency}, layer-wise profiling saves not only time but also has a smaller MSE error compared to a predictor-based strategy that is very popular in today's evolutionary-based NAS frameworks \citep{cai2019once, cai2018proxylessnas}. In addition, layer-wise profiling shows orders of magnitude better run-time when targeting hardware devices with scarce computational resources. If we consider an IoT class device as a target (i.e the Raspberry Pi Zero), it requires an unreasonably large amount of time to generate training samples for latency predictors, making them an infeasible approach in real life. For instance, the total time consumed by latency predictor is infeasible to execute on Pi Zero (last row in \Cref{tab:latency}). Of course, in reality, there is also a great number of IoT devices using more low-end CPUs compared to Pi Zero (ARMV5 or ARMV4), making the latency predictor even harder to be deployed on these devices. Also in a many-hardware setup considered in this paper, this profiling is executed O(H) times.

Most existing layer-wise look-up approaches consider at most mobile systems as targeting platforms \citep{xu2020latency,yang2018netadapt}. These systems are in general more capable than a great range of IoT devices. In this paper, we demonstrate the effectiveness of this approach on more low-end systems (Raspberry Pi and Pi Zeros), illustrating this is the more scalable approach for hardware-aware NAS for constrained hardware systems.

\paragraph{Evaluating \nas~searched architectures}
\label{sec:eval:nas}
\begin{table*}[ht!]
    \caption{
        Results of Omniglot 20-way few-shot classification.
        We keep two decimal places for our experiments, 
        and keep the decimal places as it was reported for other cited work.
        $^*$ reports a MAML replication implemented by Antoniou \etal \cite{koch2015siamese}.}
    \label{tab:omniglot}
    \begin{center}
    \adjustbox{scale=0.8}{%
    \begin{tabular}{c|cc|cc}
    \toprule
    \multirow{2}*{Method}   
    & \multirow{2}*{Size}   
    & \multirow{2}*{MACs}   
    & \multicolumn{2}{c}{Accuracy}\\
    & &        & 1-shot              & 5-shot \\
    \midrule
    Siamese Nets \cite{koch2015siamese}            
    & $35.96M$
    & $1.36G$
    & $99.2\%$            & $97.0\%$ \\
    Matching Nets \cite{vinyals2016matching}          
    & $225.91K$       
    & $20.29M$
    & $93.8\%$            & $98.5\%$ \\
    Meta-SGD \cite{li2017meta}              
    & $419.86K$      
    & $46.21M$
    & $95.93\% \pm 0.38\%$      & $98.97\% \pm 0.19\%$ \\
    \midrule
    MAML \cite{finn2017model}                        
    & $113.21K$  & $10.07M$  
    & $95.8\% \pm 0.3\%$            
    & $98.9\% \pm 0.2\%$ \\
    MAML$^*$ (Replication from \cite{antoniou2018train})                      
    & $113.21K$  & $10.07M$  
    & $91.27\% \pm 1.07\%$            
    & $98.78\%$ \\
    MAML++ $^{*}$ \cite{antoniou2018train}                    
    & $113.21K$  & $10.07M$  
    & $\mathbf{97.65\% \pm 0.05\%}$          
    & $\mathbf{99.33\% \pm 0.03\%}$ \\
    MAML++ (Local Replication)                
    & $113.21K$  & $10.07M$  
    & $96.60\% \pm 0.28\%$               
    & $99.00\% \pm 0.07\%$ \\
    \midrule
    \rowcolor{gray!15}
    H-Meta-NAS                 
    & $\mathbf{110.73K}$        
    & $\mathbf{4.95M}$
    & $97.61 \pm 0.03\%$
    & $99.11\% \pm 0.09\%$\\
    \bottomrule
    \end{tabular}
    }
    \end{center}
    \vskip -0.1in
\end{table*}
\begin{table*}[ht!]
    \caption{
        Results of Mini-ImageNet 5-way classification.
        We use two decimal places for our experiments, 
        and keep the decimal places of cited work as they were originally reported.
        T-NAS uses the complicated DARTS cell \cite{lian2019towards}, it has a smaller size but a large MACs usage.
    }
    \label{tab:miniimagenet}
    \begin{center}
    \adjustbox{scale=0.8}{%
    \begin{tabular}{c|cc|ccc}
    \toprule
    \multirow{2}*{Method}   
    & \multirow{2}*{Size}   
    & \multirow{2}*{MACs}     
    & \multicolumn{2}{c}{Accuracy}\\
                            &&        & 1-shot               & 5-shot \\
    \midrule
    Matching Nets \cite{vinyals2016matching}          
    & $228.23K$
    & $200.31M$  
    & $43.44 \pm 0.77\%$   & $55.31 \pm 0.73\%$ \\
    CompareNets \cite{sung2018learning}
    & $337.95K$
    & $318.38M$
    & $50.44 \pm 0.82\%$   & $65.32 \pm 0.70\%$ \\
    \midrule
    MAML \cite{finn2017model}
    & $\mathbf{70.09K}$  & $57.38M$  
    & $48.70 \pm 1.84\%$   & $63.11 \pm 0.92 \%$\\
    MAML++ \cite{antoniou2018train}
    & $\mathbf{70.09K}$  & $57.38M$  
    & $52.15 \pm 0.26\%$   & $68.32 \pm 0.44 \%$\\
    ALFA + MAML + L2F \cite{NEURIPS2020_ee89223a}
    & $\mathbf{70.09K}$  & $57.38M$  
    & $52.76 \pm 0.52 \%$
    & $71.44 \pm 0.45 \%$\\
    \midrule
    OFA \cite{cai2019once} (Local Replication) + MAML++
    & 82.20K
    & 33.11M
    & $51.32 \pm 0.07\%$   & $68.22 \pm 0.12 \%$ \\
    Auto-Meta \cite{kim2018auto}              
    & 98.70K
    & -
    & $51.16 \pm 0.17\%$   & $69.18 \pm 0.14 \%$ \\
    BASE (Softmax) \cite{shaw2018meta}           
    & $1200K$
    & -
    & -   
    & $65.4 \pm 0.7 \%$ \\
    BASE (Gumbel) \cite{shaw2018meta}              
    & $1200K$
    & -
    & -
    & $66.2 \pm 0.7 \%$ \\
    T-NAS $^*$ \cite{lian2019towards}                  
    & $24.3/26.5 K$  
    & $37.96/52.63 M$
    & $52.84 \pm 1.41\%$   & $67.88 \pm 0.92 \%$ \\
    T-NAS++ $^*$ \cite{lian2019towards}
    & $24.3/26.5 K$ 
    & $37.96/52.63 M$
    & $54.11 \pm 1.35\%$   & $69.59 \pm 0.85 \%$ \\
    \midrule
    \rowcolor{gray!15}
    H-Meta-NAS                     
    & $70.28K$    
    & $\mathbf{24.09M}$
    & $\mathbf{57.36 \pm 1.11 \%}$            
    & $\mathbf{77.53 \pm 0.77 \%}$    \\
    \bottomrule
    \end{tabular}
    }
    \end{center}
    \vskip -0.1in
\end{table*}

\Cref{tab:omniglot} displays the results of \nas~on the Omniglot 20-way 1-shot and 5-shot classification tasks. 
We match the size of \nas~to MAML and MAML++ for a fair comparison.
\nas~outperforms all competing methods apart from the original MAML++.
MAML++ uses a special evaluation strategy, it creates an ensemble of models with best validation-set performance. 
MAML++ then picks the best model from the ensemble based on support set loss and report accuracy on the query set.
We then locally replicated MAML++ without this trick, and show that \nas~outperforms it by a significant margin ($+1.01\%$ on 1-shot and $+0.11\%$ on 5-shot) with around half of the MACs ($4.95G$ compared to $10.07G$).

\Cref{tab:miniimagenet} shows the results of running the 5-way 1-shot and 5-shot Mini-ImageNet tasks, similar to the previous results, we match the size of searched networks to MAML, MAML++ and ALFA+MAML+L2F.
\Cref{tab:miniimagenet} not only displays results on MAML methods with fixed-architectures, it also shows the performance of searched networks including Auto-Meta \cite{kim2018auto}, BASE \cite{shaw2018meta} and T-NAS \cite{lian2019towards}.
\nas~shows interesting results when compared to T-NAS and T-NAS++. 
\nas~has a much higher accuracy ($+3.26\%$ in 1-shot and $7.94\%$ in 5-shot) and a smaller MAC count, but uses a greater amount of parameters.
T-NAS and T-NAS++ use DARTS cells \cite{liu2018darts}.
This NAS cell contains a complex routing of computational blocks, making it not suitable for latency critical applications.
We will demonstrate later how this design choice gives 
a worse on-device latency performance.
We also show how \nas~work with FC100 in Appendix.

\paragraph{\nas~for diverse hardware platforms and constraints}
\label{sec:eval:latency}
\begin{figure}[]
  \centering
    \centering
    \includegraphics[width=.5\linewidth]{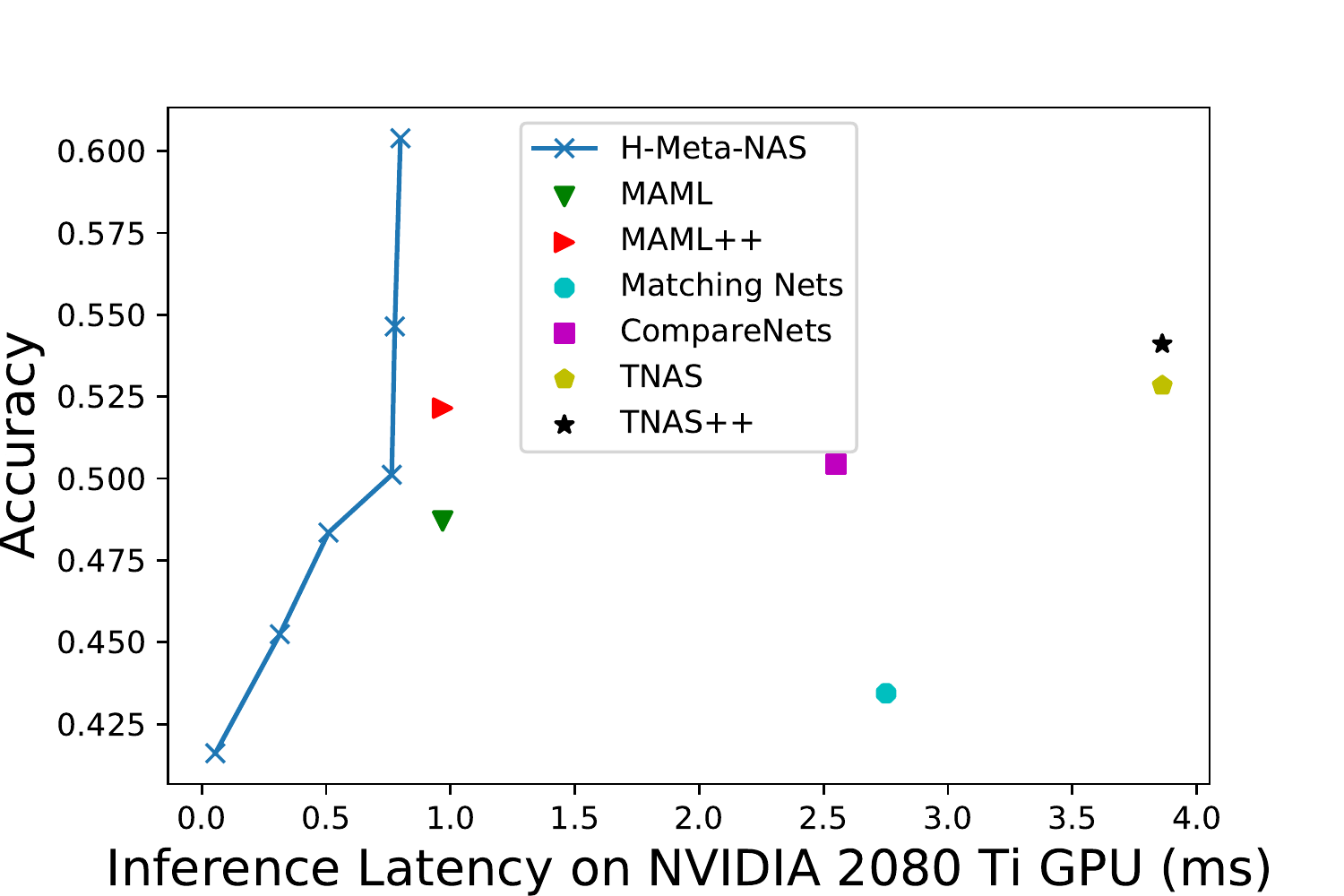}
    \caption{Targeting a GPU}
    \label{fig:gpu}
  \caption{Applying \nas~with GPU latency as optimisation targets.}
  \label{fig:sizes:gpus}
  \vspace{-15pt}
\end{figure}

In addition to using the model sizes as a constraint for \nas, we use various latency targets on various hardware platforms as the optimisation target.
\Cref{fig:sizes:gpus} shows how GPU latencies can be used as constraints.
The smaller model sizes of T-NAS do not provide a better run-time on GPU devices (\Cref{fig:gpu}), in fact, T-NAS based models have the worst run-time on GPU devices due to the complicated dependency of DARTS cells. 
We only compare to MAML and MAML++ when running on Eyeriss due to the limitations of the ScaleSIM simulator \cite{samajdar2018scale}.
In our Appendix, we provide more latency \vs~accuracy plots using various hardware platforms' latency as constraints and observe the same pareto dominance shown in \Cref{fig:sizes:gpus}.



\begin{table*}[h!]
    \caption{
        Applying H-Meta-NAS to different NAS backbones/algorithms
        for the Mini-ImageNet 5-way 1-shot classification.
    }
    \label{tab:complicate}
    \begin{center}
    \adjustbox{scale=0.8}{%
    \begin{tabular}{c|cc|ccc}
    \toprule
    Method  
    & Network Backbone
    & Inference Style
    & Size
    & MACs 
    & Accuracy\\
    \midrule
    MAML \cite{finn2017model}
    & VGG-based
    & Single Pass
    & $70.09K$  
    & $57.38M$
    & $48.70 \pm 1.84\%$   \\
    MAML++ \cite{antoniou2018train}
    & VGG-based
    & Single Pass
    & $70.09K$ 
    & $57.38M$
    & $52.15 \pm 0.26\%$   \\
    Meta-Baseline \cite{chen2020new}
    & ResNet-based
    & Multi Pass
    & $12.44M$ 
    & $56.48G$
    & $63.17 \pm 0.23\%$   \\
    DeepEMD \cite{zhang2020deepemd}
    & ResNet-based
    & Multi Pass
    & $12.44M$
    & $56.38G$
    & $65.91 \pm 0.82\%$   \\
    \midrule
    \rowcolor{gray!15}
    H-Meta-NAS
    & VGG-based
    & Single Pass
    & $70.28K$    
    & $24.09M$
    & $57.36 \pm 1.11 \%$ \\
    \rowcolor{gray!15}
    H-Meta-NAS
    & ResNet-based
    & Single Pass
    & $70.62K$    
    & $28.19M$
    & $64.67 \pm 2.03 \%$ \\
    \bottomrule
    \end{tabular}
    }
    \end{center}
    \vskip -0.1in
\end{table*}  
\begin{table*}[h!]
    \caption{
        Comparing the NAS search complexity with $N$ tasks, $H$ hardware platforms and $C$ constraints.
        Search time is estimated for a deployment scenario with 500 tasks and 10 hardware-constraint pairs, estimation details are discussed in Appendix.}
        
    \label{tab:searchtime}
    \begin{center}
    \adjustbox{scale=.8}{%
    \begin{tabular}{c|ccccc}
    \toprule
    Method  
    & Style 
    & Hardware-aware
    & Search complexity
    & Search time (GPU hrs)
    \\
    \midrule
    DARTS \cite{liu2018darts}
    & Gradient-based, single task 
    & No
    & $\mathcal{O}(THC)$
    & $\approx 10^6$
    \\
    Once-for-all \cite{cai2019once}
    & Evolution-based, single task 
    & Yes
    & $\mathcal{O}(N)$
    & $\approx 10^4$
    \\
    TNAS \& TNAS++ \cite{lian2019towards}
    & Gradient-based, multi task 
    & No
    & $\mathcal{O}(HC)$
    & $\approx 10^3$
    \\
    \rowcolor{gray!15}
    \nas 
    & Evolution-based, multi task 
    & Yes 
    & $\mathcal{O}(1)$  
    & 40
    \\
    \bottomrule
    \end{tabular}
    }
    \end{center}
    \vskip -0.1in
\end{table*}

\paragraph{A more complex NAS backbone}
\Cref{tab:complicate} shows how \nas~performs with a more complicated NAS backbone.
In previous experiments, we build the NAS on top of a VGG9 backbone since it is the architecture utilised in the MAML++ algorithm.
For the purpose of having a fair comparison, we did not manually pick a complex NAS backbone.
However, we demonstrate, in this section, that \nas~can be applied with a more complicated backbone and it shows better final accuracy as expected.
The trained accuracy of searched networks using ResNet12 reaches a $7.31\%$ higher accuracy compared the original VGG9 backbone.
In addition, we compare the proposed approach with state-of-the-art Metric-based meta-learning methods \cite{zhang2020deepemd,chen2020new}.
Although using only a single inference pass (our method does not conduct inference runs on the support set when deployed), \nas~shows competitive results with SOTA Metric-based methods while having a much smaller MACs usage (around $20 \times$).
\paragraph{Search complexity and search time}
In \Cref{tab:searchtime}, we show a comparison between \nas~and various NAS schemes in the many-task many-device setup. Specifically, we consider a scenario with $500$ tasks and 10 different hardware-constraint paris.
Our results in \Cref{tab:searchtime} suggest that \nas~is the most efficient search method because of its low search complexity.


\section{Conclusion}
In this paper, we show \nas, a NAS method focusing on fast adaption of not only model weights but also model architectures in a many-task many-device few-shot learning setup.
\nas~shows a Pareto dominance when compared to a wide range of MAML baselines and other NAS results.
We study the effectiveness of \nas~on a wide variety of hardware systems and constraints, and demonstrate its superior performance on real-hardware devices using an orders of magnitude shorter search time compared to existing NAS methods.
\newpage
\appendix

\section{Details of VGG9 and ResNet12 backbones}
\Cref{tab:backbone:vgg} and \Cref{tab:backbone:resnet12} show the NAS backbones of \nas. Clearly the ResNet-based NAS backbone is significantly more complicated.
The kernel size search space is $\{1, 3, 5\}$. 
The channel expansion search space is $\{0.25, 0.5, 0.75, 1, 1.5, 2, 2.25\}$ for the VGG-based NAS backbone but $\{0.25, 0.5, 1, 1.5, 1.75, 2\}$ for the ResNet-based backbone.
The reason for the modification in search space is because the GPU RAM limitation does not support an expansion size of $2.25$ on the ResNet-based backbone.
The activation search space contains 
$\{['relu', 'elu', 'selu', 'sigmoid', 'relu6', 'leakyrelu'\}$.

\begin{table}[h!]
    \caption{
      Details of the VGG9 NAS backbone}
    \label{tab:backbone:vgg}
    \vskip 0.15in
    \begin{center}
    \adjustbox{max width=1.\textwidth}{%
    \begin{tabular}{ccc}
      \toprule
      Layer Name
      & Base channel counts 
      & Stride
      \\
      \midrule
      Layer0 
      & 64 
      & 2
      \\
      Layer1
      & 64 
      & 2
      \\
      Layer2
      & 64 
      & 2
      \\
      Layer3
      & 64 
      & 2
      \\
      \bottomrule
      \end{tabular}
  }
    \end{center}
    \vskip -0.1in
  \end{table}

  \begin{table}[h!]
    \caption{
      Details of the ResNet12 NAS backbone}
    \label{tab:backbone:resnet12}
    \vskip 0.15in
    \begin{center}
    \adjustbox{max width=1.\textwidth}{%
    \begin{tabular}{ccc}
      \toprule
      Layer Name
      & Base channel counts 
      & Stride
      \\
      \midrule
      Block0\_Layer0 
      & 32 
      & 2 \\
      Block0\_Layer1
      & 32 
      & 1 \\
      Block0\_Layer2
      & 32 
      & 1
      \\
      Block1\_Layer0 
      & 64
      & 2 \\
      Block1\_Layer1
      & 64 
      & 1 \\
      Block1\_Layer2
      & 64 
      & 1 \\
      Block2\_Layer0 
      & 128
      & 2 \\
      Block2\_Layer1
      & 128
      & 1 \\
      Block2\_Layer2
      & 128
      & 1 \\
      Block3\_Layer0 
      & 256
      & 2 \\
      Block3\_Layer1
      & 256
      & 1 \\
      Block3\_Layer2
      & 256
      & 1 \\
      \bottomrule
      \end{tabular}
  }
    \end{center}
    \vskip -0.1in
\end{table}

\section{Tuning the decay process in pre-training strategy}
As mentioned in Section 3.3 in the paper, we apply a progressive shrinking strategy to pre-training. We decay the probability of picking the largest sub-network gradually.
Recall that the architectural sampling process $\alpha \sim p(\mathbb{A})$ will pick the largest network with a probability of $p$, and randomly pick a sub-network with a probability of $1-p$.
We apply an exponentially decay strategy to $p$:

\begin{equation}
	p = p_e + (p_i - p_e) \times exp(- \alpha \times \frac{e-e_s}{e_{m} - e_s}))
\end{equation}
$p_e$ and $p_i$ are the end and initial probabilities. $e$ is the current number of epochs, and $e_s$ is the starting epoch of applying this decaying process. $\alpha$ determines how fast the decay is.
In our experiment, we pick $p_i=1.0$ and $e_s=30$, because the super-net reaches a relatively stable training accuracy at that point.
We then start the decaying process, and evaluate different values of $\alpha$ in \Cref{tab:alpha}.
The averaged accuracy is averaged across 100 randomly picked sub-networks on the $\mathcal{T}_{val}$ tasks.
Based on these results, we picked $\alpha=5$ for our later experiments.
\begin{table}[h!]
    \caption{
        Tuning the decay factor $\alpha$ for pre-training on Mini-ImageNet 5-way 1-shot classification.
		Accuracy is averaged across 100 randomly picked sub-networks.
    }
    \label{tab:alpha}
    \vskip 0.15in
    \begin{center}
    \begin{small}
    \begin{sc}
    \begin{tabular}{c|ccccc}
    \toprule
    $\alpha$	
	& 0.1		& 0.5	 & 5 		& 10		& 50 \\
    Avg Accuracy	
	& 0.424		& 0.4145 & 0.5464 	& 0.5323	& 0.4423\\
    \bottomrule
    \end{tabular}
    \end{sc}
    \end{small}
    \end{center}
    \vskip -0.1in
\end{table}

\section{Adaption algorithm and the hyper-parameter choices}
\Cref{alg:adapt} details the adaption algorithm. In the $Mutate$ function, each architecture is ranked with the averaged loss across all sampled tasks, and $10\%$ of the architectures with the lowest loss values are then used to perform a classic genetic algorithm mutation \cite{whitley1994genetic}.
The mutation will allow the top-performing architectures to have two randomly picked architectural choices being modified to another choice that is not the original one.
The mutation function considers the original pool of architectures ($\mathbb{A}$) and their averaged loss values ($L_a$). 
The cost of each architecture can be computed by the pre-build hardware-specific hash-table $H_t(\mathcal{A})$.
We then only mutate the subset in $\mathbb{A}$ that their hardware cost has satisfied the constraints $\{\mathcal{A} | \mathcal{A} \in \mathbb{A} \wedge H_t(\mathcal{A}) \leq C\}$.
The mutation is to randomly pick two options in the entire architectural space and change them to other choices that are different from the original.


\begin{algorithm}[!h]
	\caption{The adaption algorithm}
	\label{alg:adapt}
	\begin{algorithmic}
		\State{\bfseries Input:} $M$, $P$, $C$, $H_t$ 
		\State{$\mathbb{A}=Init(P)$} 
    \algorithmiccomment{Initialise a set of architectures with a size of $P$}
		\For{$i=0$ {\bfseries to} $M - 1$}
      \State{$L_{a} = \emptyset$}
      \State{$\mathbb{T}_s \sim p(\mathbb{T}_{val})$}
      \algorithmiccomment{Obtain a subset from the validation task set}
      \For{$\mathcal{A} \in \mathbb{A}$}
        \State{$L_{t} = \emptyset$}
        \For{$\mathcal{T} \in \mathbb{T}_s$}
          \State{$l = \mathcal{L}(\mathcal{T}, \mathcal{A})$} \algorithmiccomment{Compute loss}
          \State{$L_{t} = L_{t} \bigcup \{l\}$}
        \EndFor{}
        \State{$L_a = L_{a} \bigcup \{mean(L_{t})\}$} \algorithmiccomment{Collect averaged loss values across all tasks}
      \EndFor{}
    \State{$\mathbb{A} = Mutate(\mathbb{A}, L_a, H_t, C)$} \algorithmiccomment{Mutate the architectures based on hardware constraints}

		\EndFor{}
	\end{algorithmic}
\end{algorithm}



\section{\nas~search configurations and hardware simulation}
We mostly follow the experiment setup in MAML++ \cite{antoniou2018train}.
In the pre-training stage, we train for 100 epochs, each epoch consists of 500 iterations.
We also pick 600 tasks to be validation tasks. 
In the adaption stage, we randomly sample from the validation set, and pick 16 tasks to build a data slice for the architectures to traverse.
In the final re-training stage of a searched architecture, we follow the strategy used in MAML++ \cite{antoniou2018train}.
We then introduce the detailed special configurations for the datasets:
\begin{itemize}
  \item Omniglot: We randomly split 1200 characters for training, and the rest is used for testing. The images are augmented with randomised rotation of multiples of 90 degrees.
  \item Mini-ImageNet: All images are down-sampled to $84 \times 84$.
\end{itemize}

We use the ScaleSim framework \cite{samajdar2018scale} for simulating the Eyeriss \cite{chen2016eyeriss} accelerator.
ScaleSim is an open-source cycle-accurate CNN simulator. The simulator has certain limitations with respect to the DRAM simulation, it could be advanced with an external DRAM simulator but will cause a large run-time. So we kept the original setup and the DRAM simulation would report a read/write bandwidth requirements. For simplicity, we assume these DRAM requirements are met.
In addition, it is a well-known fact that cycle-accurate simulators are slow to execute. Due to this reason, we only launched the MAML and MAML++ networks in the ScaleSim simulator.

\section{Additional results on FC100}
In \Cref{tab:fc100}, we further demonstrate the effectiveness of the proposed \nas~on the FC100 dataset.
T-NAS did not report their model sizes on this task, and our results suggest that \nas~achieves the best accuracy on both the 1-shot and 5-shot setups.
\begin{table}[h!]
    \caption{
        Results of FC100 5-way few-shot classification.
        We keep two decimal places for our experiments, 
        and keep the decimal places of cited work as they were originally reported.
    }
    \label{tab:fc100}
    \vskip 0.15in
    \begin{center}
    \begin{small}
    \begin{sc}
    \begin{tabular}{c|c|cc}
    \toprule
    \multirow{2}*{Method}   & \multirow{2}*{Size}   & \multicolumn{2}{c}{Accuracy}\\
                            &        & 1-shot              & 5-shot \\
    \midrule
    MAML                    
    & $70.09 K$       
    & $38.1 \pm 1.7\%$    & $50.4 \pm 1.0 \%$ \\
    MAML++                  
    & $70.09K$        
    & $38.7 \pm 0.4\%$    & $52.9 \pm 0.4 \%$\\
    \midrule
    T-NAS                   & -        & $39.7 \pm 1.4\%$    & $53.1 \pm 1.0 \%$ \\
    T-NAS++                 & -       & $40.4 \pm 1.2\%$    & $54.6 \pm 0.9 \%$ \\
    \midrule
    H-META-NAS               
    & $55.52K$  
    & $43.29 \pm 1.22 \%$             
    & $56.86 \pm 0.76 \%$ \\
    \bottomrule
    \end{tabular}
    \end{sc}
    \end{small}
    \end{center}
    \vskip -0.1in
\end{table}

\section{T-NAS baseline results}
We notice the model sizes of some baseline models (\eg~MAML and MAML++) reported in the original TNAS paper \cite{lian2019towards} are different from our results in Table 3.
We calculated the model sizes of these baselines using their official open-sourced implementations.
T-NAS did not provide an implementation of their mentioned baselines in their official repository, 
so we cannot replicate their model size numbers. We have contacted the T-NAS authors regarding this issue.

\section{Search time estimation}
Due to the limited computing facilities available, we estimate the search time of DARTS \cite{liu2018darts}, Once-for-all \cite{cai2019once} and T-NAS \cite{lian2019towards} in a multi-task multi-device setup.
We take the search time reported in the original publications and multiply them by the appropriate scaling factors.
For DARTs, we take the search time (4 GPU days = 96 GPU hours) and multiply it by $H \times T = 5000$.
We additionally assume a linear scaling relationship between search time and input image sizes, so we multiply the total search time by $\frac{84 \times 84}{32 \times 32}$, this gives us in total a search time of around $ 10^6$.
We perform the same estimation for Once-for-all \cite{cai2019once} and T-NAS \cite{lian2019towards}.

\section{Latency-aware optimisation on more hardware platforms}
\begin{figure}[!h]
  \centering
    \centering
    \includegraphics[width=.5\linewidth]{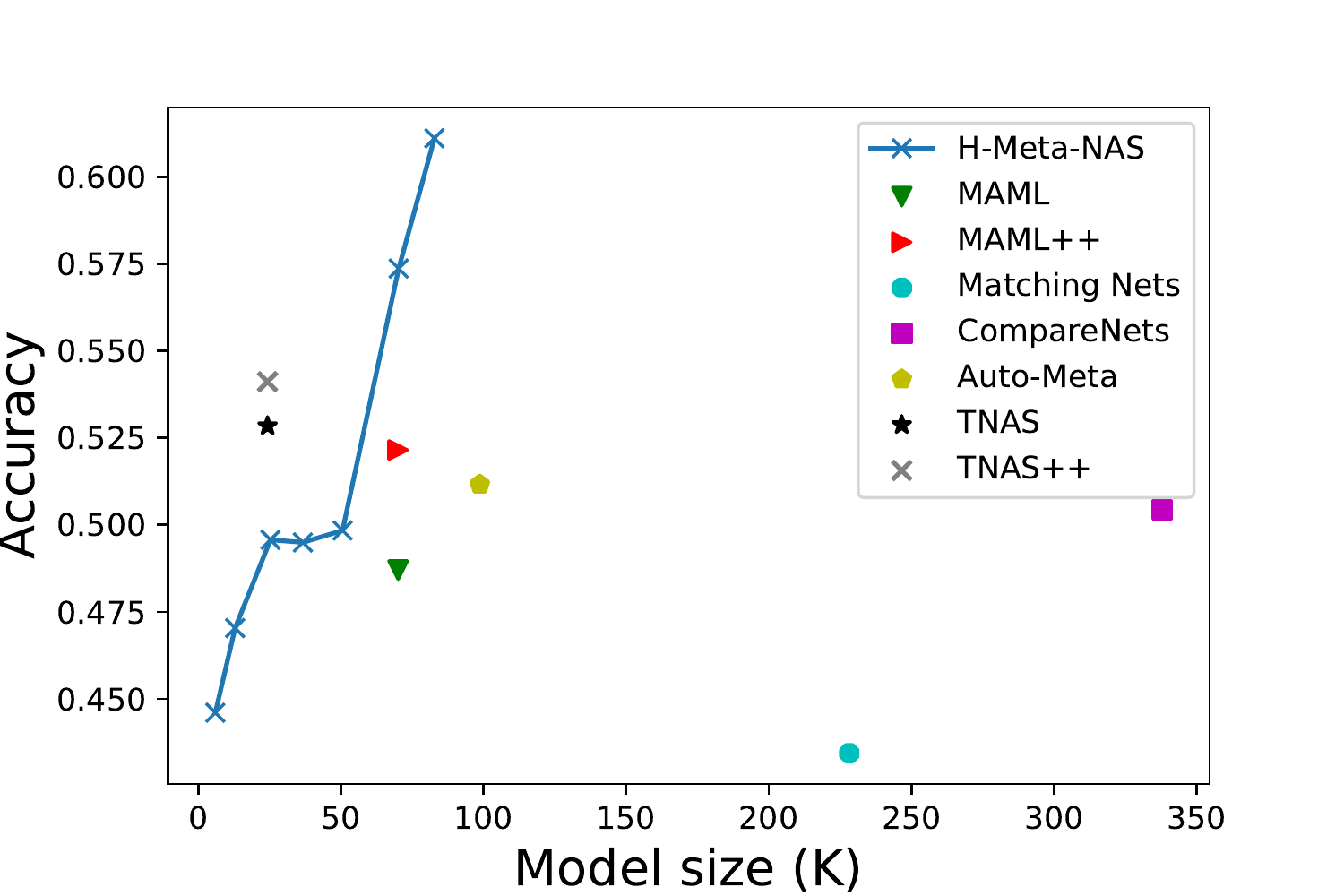}
    \caption{Targeting model sizes}
    \label{fig:gpu}
  \caption{Applying \nas~with model sizes as optimisation targets.}
  \label{fig:sizes}
\end{figure}

\begin{figure}[!h]
    \centering
    \begin{subfigure}{.5\textwidth}
      \centering
      \includegraphics[width=.95\linewidth]{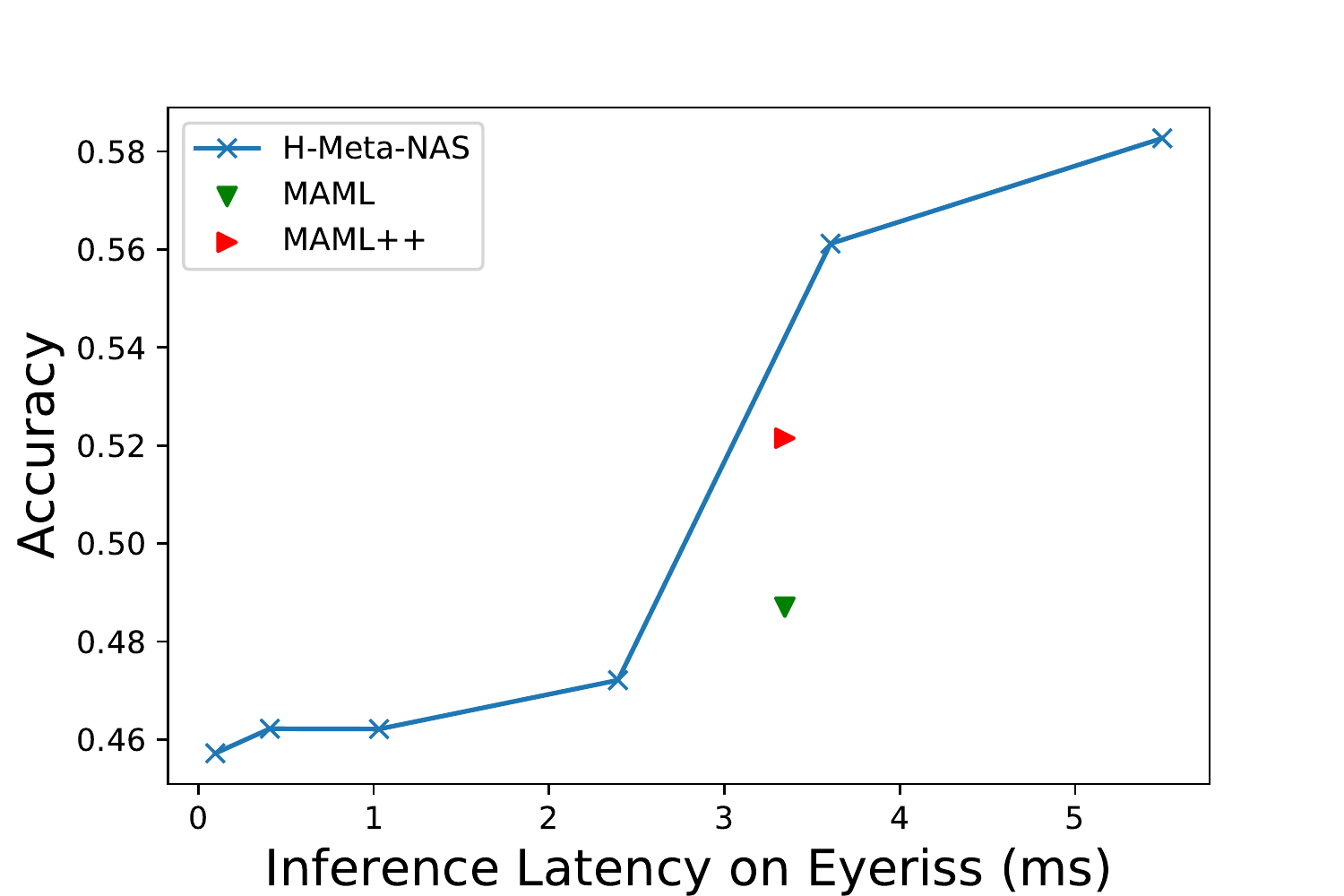}
      \caption{Targeting an ASIC accelerator}
      \label{fig:asic}
    \end{subfigure}%
    \begin{subfigure}{.5\textwidth}
      \centering
      \includegraphics[width=.95\linewidth]{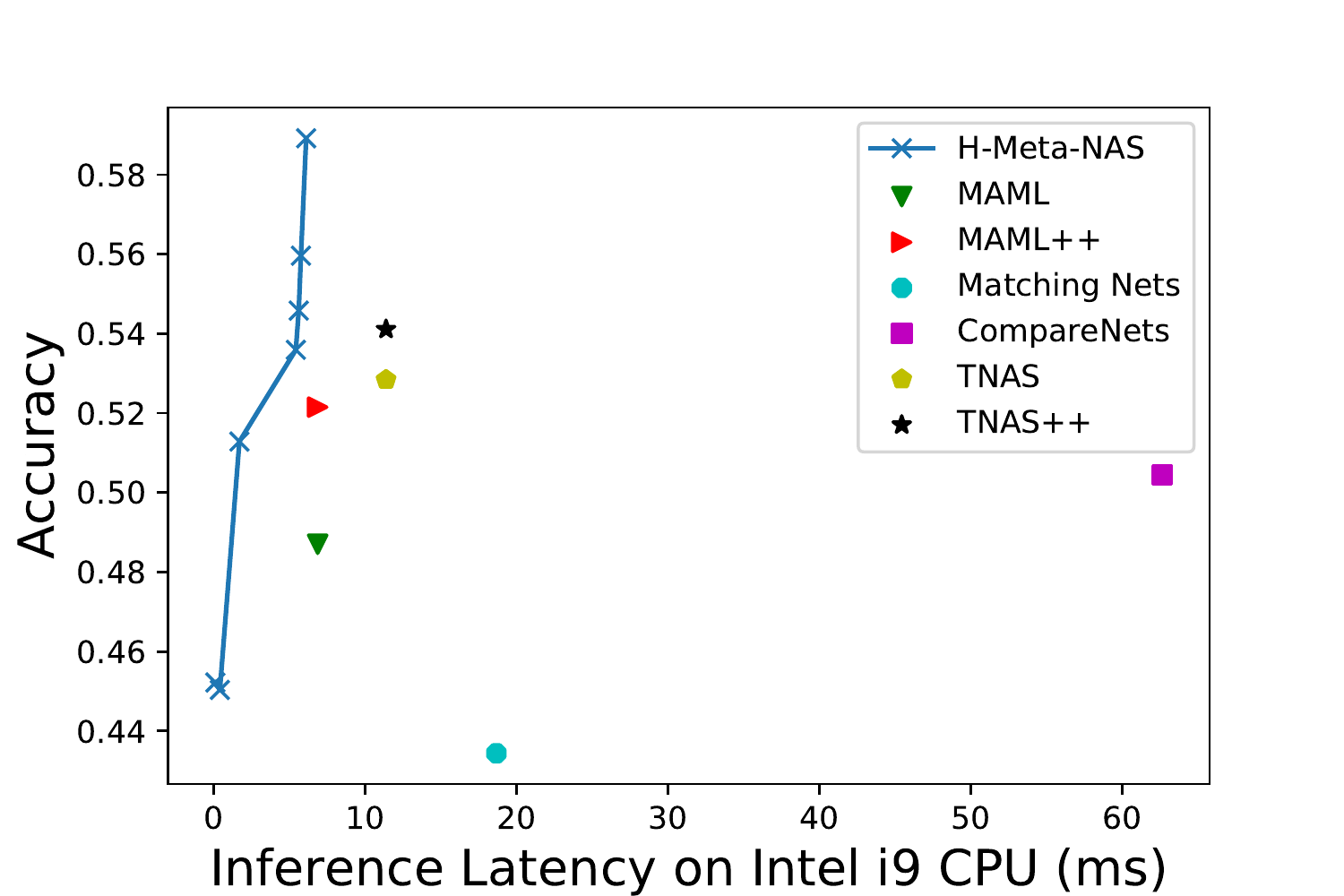}
      \caption{Targeting a mid-end CPU}
      \label{fig:midcpu}
    \end{subfigure}

    \vskip\baselineskip
    \begin{subfigure}{.5\textwidth}
      \centering
      \includegraphics[width=.95\linewidth]{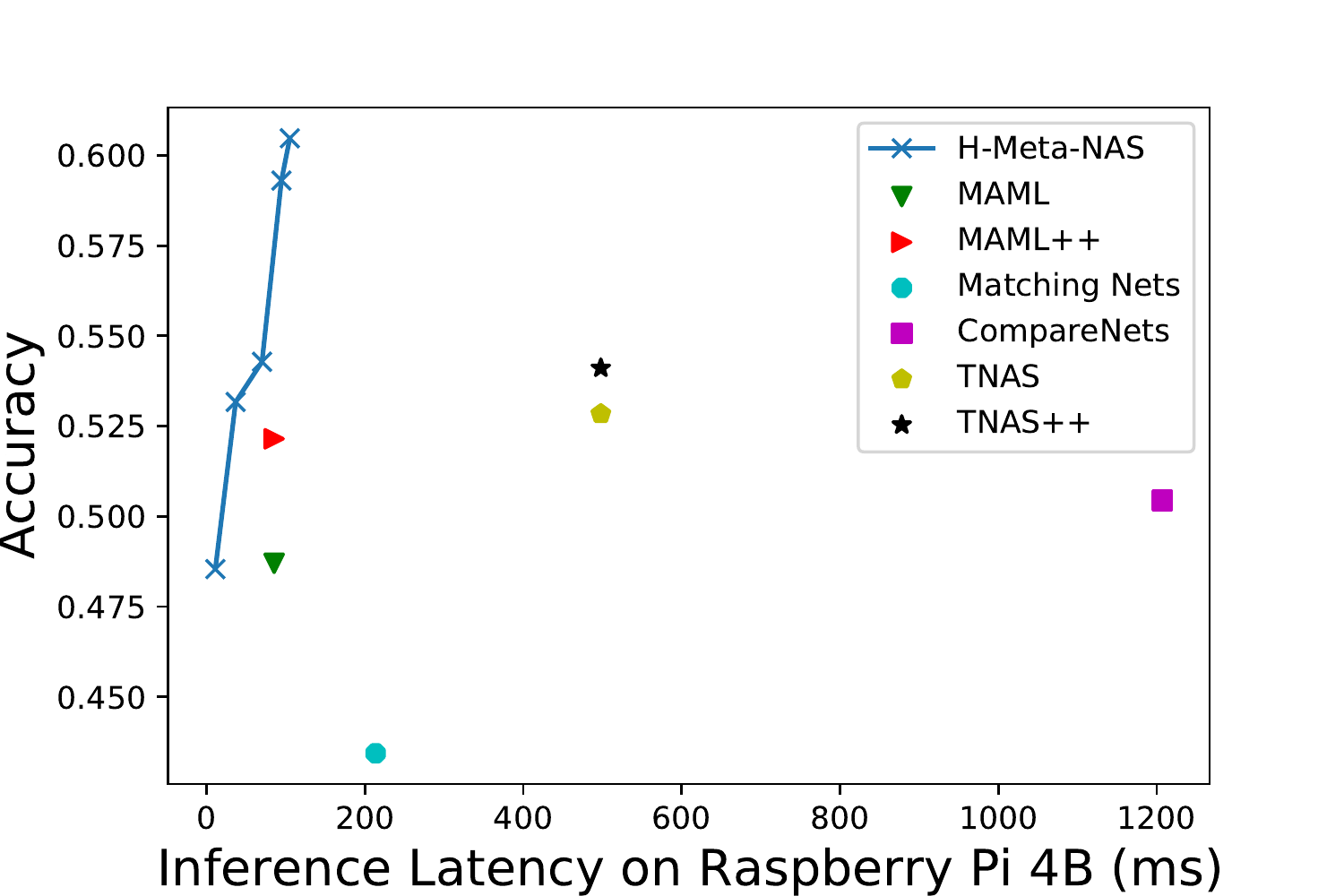}
      \caption{Targeting a low-end CPU} 
      \label{fig:lowcpu}
    \end{subfigure}%
    \begin{subfigure}{.5\textwidth}
      \centering
      \includegraphics[width=.95\linewidth]{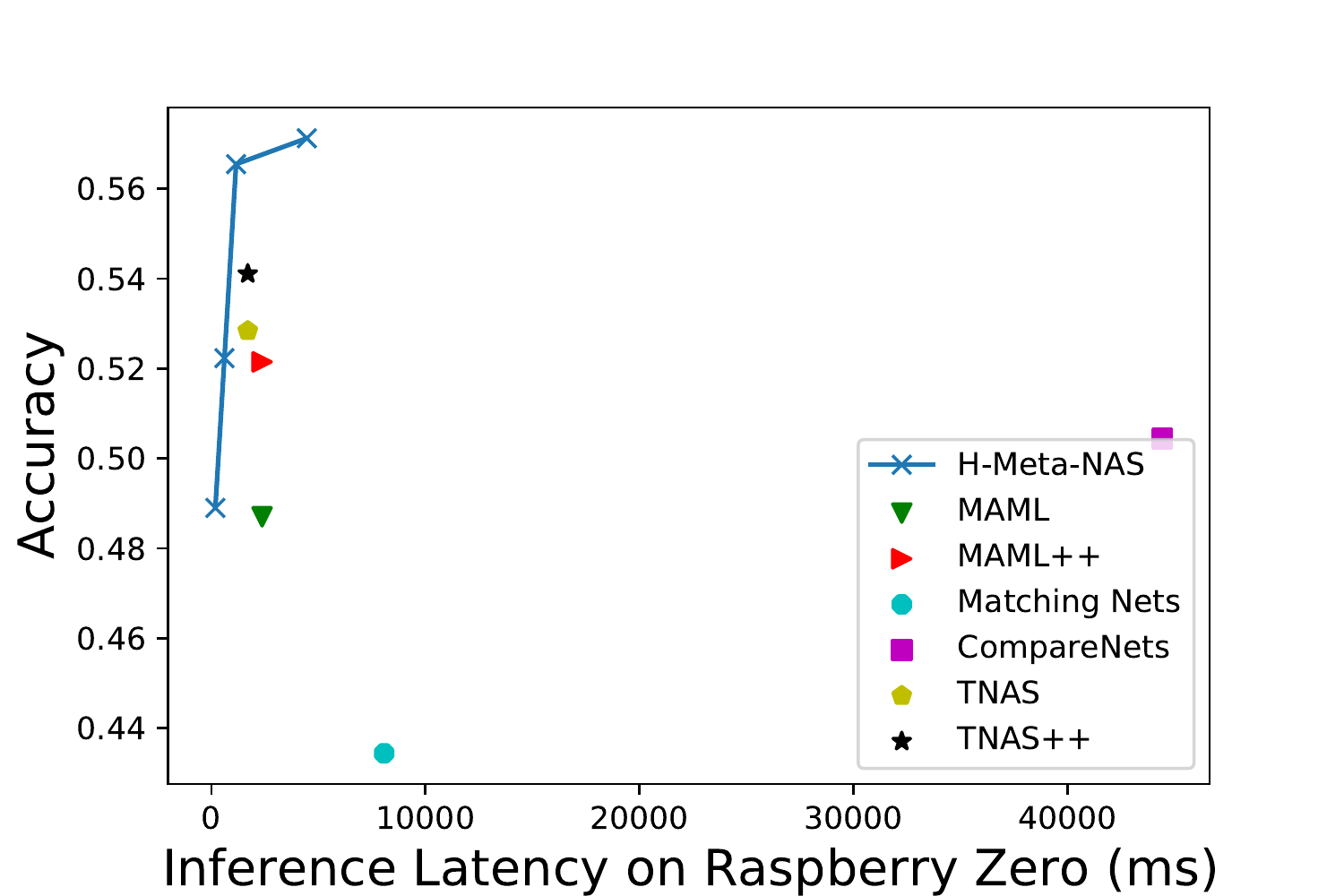}
      \caption{Targeting an IoT device}
      \label{fig:iot}
    \end{subfigure}
    \caption{Applying \nas~with ASIC and CPUs as targets.}
    \label{fig:cpu:asic:iot}
\end{figure}

In addition to using the model sizes as a constraint for \nas, we use various latency targets on various hardware platforms as the optimisation target.
\Cref{fig:sizes} shows how GPU latencies can be used as constraints.
T-NAS and T-NAS++ show a better performance on the size-accuracy plot in \Cref{fig:sizes}.
The smaller model sizes of T-NAS do not provide a better run-time on GPU devices (\Cref{fig:gpu}), in fact, T-NAS based models have the worst run-time on GPU devices due to the complicated dependency of DARTS cells. 
\Cref{fig:cpu:asic:iot} illustrates the performance of \nas~on different CPU devices and an ASIC hardware. The details of these hardware are described in the main paper.
In \Cref{fig:cpu:asic:iot}, \nas~shows a better Pareto-frontier performance compared to a range of baselines and searched models.
We only compare to MAML and MAML++ when running on Eyeriss due to the limitations of the ScaleSIM simulator \cite{samajdar2018scale}.
In our Appendix, we provide more latency \vs~accuracy plots using various hardware platforms' latency as constraints and observe the same pareto dominance shown in \Cref{fig:cpu:asic:iot}.
Our results in \Cref{fig:cpu:asic:iot} demonstrate that \nas~consistently generates more efficient models compared to various MAML-based methods.

\section{License of the assets}
In our work, we utilised the following datasets/library/code:

\begin{table}[h!]
  \caption{
    Licenses of used assets.
  }
  \label{tab:license}
  \vskip 0.15in
  \begin{center}
  \begin{small}
  \begin{sc}
  \begin{tabular}{c|c}
  \toprule
  Dataset/algorithm/lib names & License \\
  \midrule
  The Omniglot Dataset & MIT License \\
  The Mini-ImageNet Dataset & MIT License \\
  The FC100 Dataset &  Apache V2 License \\
  Pytorch-Meta &  MIT License \\
  MAML++ &  MIT License \\
  \bottomrule
  \end{tabular}
  \end{sc}
  \end{small}
  \end{center}
  \vskip -0.1in
\end{table}
\newpage
\bibliographystyle{unsrtnat}
\bibliography{references}  






\end{document}